\documentclass[journal]{IEEEtran}

\ifCLASSINFOpdf
\else
   \usepackage[dvips]{graphicx}
\fi
\usepackage{url}

\usepackage{graphicx}
\usepackage{cite}
\usepackage{epsfig}
\usepackage{amsmath}
\usepackage{amssymb}
\usepackage{comment}
\usepackage{booktabs}
\usepackage{multirow}
\usepackage{{stfloats}}

\usepackage{array, caption, threeparttable}
\usepackage{caption}

\renewcommand\appendix{\setcounter{secnumdepth}{-2}}

\begin{document}

\title{TransRPPG: Remote Photoplethysmography Transformer for 3D Mask Face Presentation \\Attack Detection}

\author{Zitong Yu, Xiaobai Li, Pichao Wang and Guoying Zhao, \IEEEmembership{Senior Member, IEEE}
\thanks{This work was supported by the Academy of Finland for project MiGA (grant 316765), ICT 2023 project (grant 328115), Infotech Oulu, project 6+E (grant 323287) funded by Academy of Finland, and project PhInGAIN (grant 200414) funded by The Finnish Work Environmental Fund. The authors wish to acknowledge CSC-IT Center for Science, Finland, for computational resources. (Corresponding author: Guoying Zhao)}
\thanks{Z. Yu, X. Li and G. Zhao are with the Center for Machine Vision and Signal
Analysis, University of Oulu, Oulu 90014, Finland. E-mail: \{zitong.yu, xiaobai.li,  guoying.zhao\}@oulu.fi.}
\thanks{P. Wang is with Alibaba Group, Bellevue,
WA, 98004, USA. E-mail: pichao.wang@alibaba-inc.com}}

%\markboth{IEEE SIGNAL PROCESSING LETTERS}
%{Shell \MakeLowercase{\textit{et al.}}: Bare Demo of IEEEtran.cls for IEEE Journals}
\maketitle

\begin{abstract}
3D mask face presentation attack detection (PAD) plays a vital role in securing face recognition systems from the emergent 3D mask attacks. Recently, remote photoplethysmography (rPPG) has been developed as an intrinsic liveness clue for 3D mask PAD without relying on the mask appearance. However, the rPPG features for 3D mask PAD are still needed expert knowledge to design manually, which limits its further progress in the deep learning and big data era. In this letter, we propose a pure rPPG transformer (TransRPPG) framework for learning intrinsic liveness representation efficiently. At first, rPPG-based multi-scale spatial-temporal maps (MSTmap) are constructed from facial skin and background regions. Then the transformer fully mines the global relationship within MSTmaps for liveness representation, and gives a binary prediction for 3D mask detection. Comprehensive experiments are 
conducted on two benchmark datasets to demonstrate the efficacy of the TransRPPG on both intra- and cross-dataset testings. Our TransRPPG is lightweight and efficient (with only 547K parameters and 763M FLOPs), which is promising for mobile-level applications.

\end{abstract}


\IEEEpeerreviewmaketitle

\vspace{-0.4em}
\section{Introduction}

\IEEEPARstart{F}{ace} recognition~\cite{guo2020learning} technology has been widely used in many interactive intelligent systems due to their convenience and remarkable accuracy. However, face recognition systems are still vulnerable to presentation attacks (PAs) ranging from print~\cite{zhang2012face}, replay~\cite{li2019replayed} and emerging 3D mask~\cite{liu20163d} attacks. Therefore, both the academia and industry have realized the critical role of face presentation attack detection (PAD)~\cite{liu2021cross} technique for securing the face recognition system.

In the past decade, both traditional~\cite{wen2015face,boulkenafet2016face22,Patel2016Secure} and deep learning-based~\cite{yu2020searching,yu2020face,Liu2018Learning,jourabloo2018face,yu2020fas,yu2021revisiting} methods have shown effectiveness for face PAD especially for 2D (e.g., print and replay) attacks. On one hand, most previous approaches extract the static appearance based clues (e.g., color texture~\cite{boulkenafet2016face} and noise artifacts~\cite{jourabloo2018face}) to distinguish the PAs from the bonafide. On the other hand, a few methods exploit the dynamic inconsistency (e.g., motion blur~\cite{li2019replayed} and temporal depth~\cite{wang2020deep}) between the bonafide and PAs for discrimination. Benefited from the above-mentioned advanced techniques, 2D PA detection has made satisfied progress under most normal scenarios.

Meanwhile, 3D mask attack has attracted increasing attention since the customized
3D mask can be easily made at an affordable price~\cite{erdogmus2014spoofing}. Despite strong ability on detecting low-quality 3D print mask, appearance-based methods still suffer from performance drops when encountering high-fidelity 3D masks with fine-grained texture and shape as real faces~\cite{jia2020survey}. In contrast, some researchers devote to exploring the feasibility to utilize appearance-independent liveness clues (e.g., remote photoplethysmography (rPPG)~\cite{li2016generalized,lin2019face,liu20163d22,liu2018remote}) for 3D mask PAD instead of the traditional texture/motion-based patterns.

rPPG~\cite{chen2018video} is a new technique to recover the physiological signals under ambient light~\cite{verkruysse2008remote} via analyzing the skin color changes caused by periodic cardiac heartbeats. In the early stage, most methods~\cite{verkruysse2008remote,poh2010advancements,li2014remote,de2013robust,wang2016algorithmic} capture subtle color changes on particular facial regions of interest (ROI). Then some end-to-end deep learning methods~\cite{vspetlik2018visual,chen2018deepphys,yu2019remote1,yu2019remote2,yu2020autohr} are proposed to recover rPPG signals from facial videos directly. Recently, learning upon the constructed spatio-temporal map (STmap)~\cite{niu2019rhythmnet,niu2020video} consisting of raw color variance signals from multiple facial ROIs, are proven to learn rPPG features more efficiently.

One nature question is that why rPPG is suitable for 3D mask face PAD? On one hand, rPPG can be captured using only a simple RGB camera under ambient light, which satisfies most of the video recording conditions in 3D mask face PAD. On the other hand, live/periodic pulse signals can only be observed on genuine faces but not on masked faces because the 3D mask blocks the light transmission from the facial skin~\cite{li2016generalized}. As the the rPPG clues are independent to the mask appearance, the rPPG-based 3D mask PAD methods~\cite{liu20163d22,liu2018remote,li2016generalized} can detect the high-fidelity mask well and shows good generalization capacity.

\begin{figure*}
\centering
\includegraphics[scale=0.36]{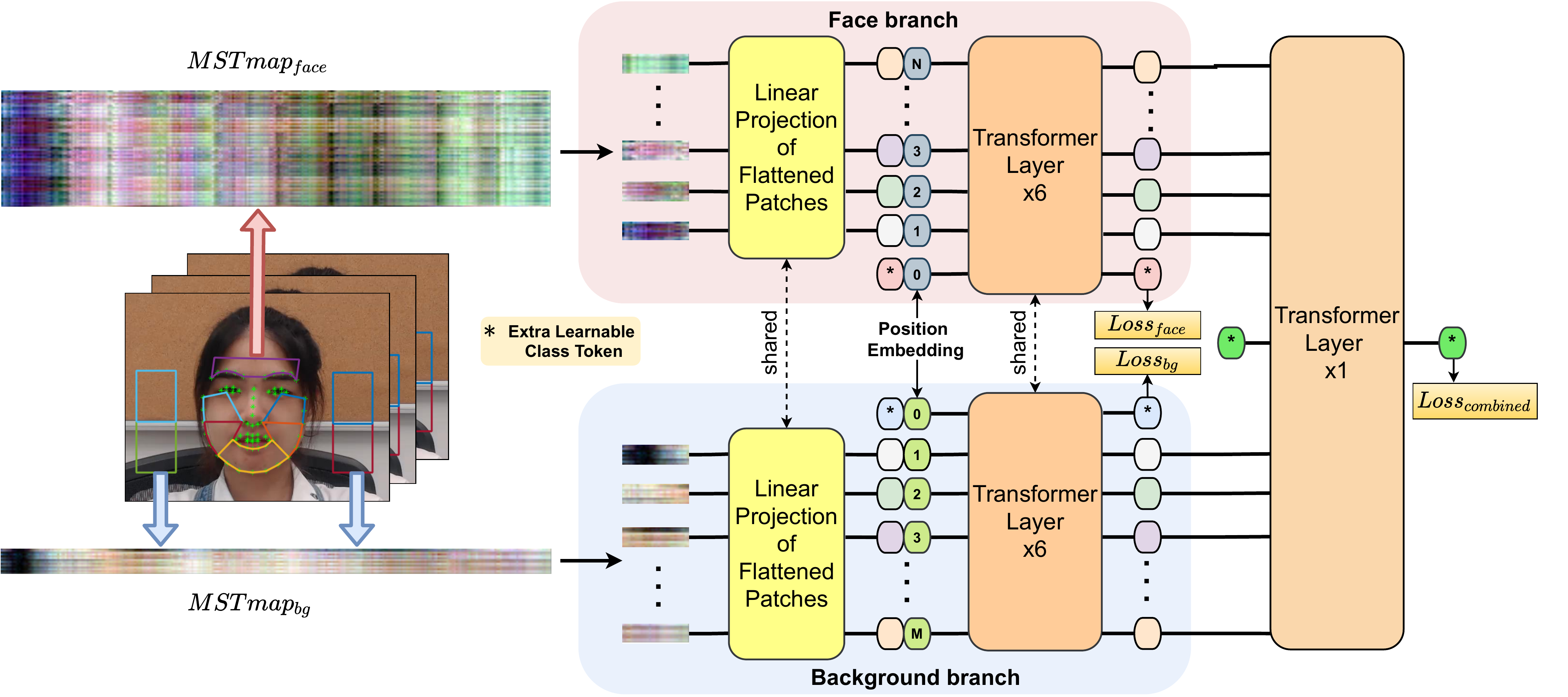}
\vspace{-0.6em}
\caption{\small{Overview of TransRPPG for 3D mask face PAD. Given a face video, two MSTmaps are constructed from facial and background region, respectively. Then two-branch vision transformers (shared) are used to extract the respective rPPG and environmental features. Finally, the combined features are refined via an extra transformer for binary (bonafide or mask attack) prediction. }}
\label{fig:framework}
\vspace{-0.6em}
\end{figure*}

To alleviate the influence of environmental noise, existing 3D mask PAD approaches~\cite{liu2018remote,li2016generalized} usually design complex hand-crafted rPPG features
(e.g., cross-correlation spectrum). In other words, how to efficiently represent rPPG features for 3D mask face PAD in a data-driven fashion is still unexplored. Recently, transformer~\cite{liu2018remote} shows its strong long-range relationship modeling ability, and achieves state-of-the-art performance in both natural language processing (NLP)~\cite{devlin2018bert} and computer vision (CV)~\cite{carion2020end,dosovitskiy2020image,he2021transreid} tasks. Inspired by the rich global attention characteristic of the vision transformer (ViT)~\cite{dosovitskiy2020image}, in this letter we propose a pure rPPG transformer framework, named TransRPPG, for learning intrinsic liveness representation efficiently. Similar to the work~\cite{niu2020video}, we first construct rPPG-based multi-scale spatial-temporal maps (MSTmap) on facial skin and background regions as the input of vision transformer~\cite{dosovitskiy2020image}. Then the transformer fully mines the global relationship within MSTmap for liveness representation, and gives a binary prediction for 3D mask detection. Our contribution includes:

\begin{itemize}
\setlength\itemsep{-0.1em}
    \item We propose the first pure transformer framework (TranRPPG) for automatic rPPG feature representation on 3D mask face PAD task. We also explore various task-aware inputs, architectures, and loss functions for TranRPPG, which is insightful for both rPPG and 3D mask face PAD communities.

    \item We conduct intra- and cross-dataset tests and show that the TranRPPG achieves superior or on par state-of-the-art performance on 3D mask face PAD. Moreover, our TransRPPG is lightweight and efficient (with only 547K parameters and 763M FLOPs), which is promising for mobile-level applications.
\end{itemize}

%\vspace{-1em}
\section{Methodology}
%\vspace{-0.5em}

In this section, we first give details about the multi-scale spatial-temporal map generation. Then we introduce the transformer-based backbone for 3D mask face PAD. An overview of the proposed method is illustrated in Fig.~\ref{fig:framework}.

\subsection{Multi-scale Spatial Temporal Map Generation}
\label{MSTmap}

To explicitly utilize sufficient rPPG signals from global and local regions for 3D mask face PAD, we follow the similar manner about multi-scale spatial temporal maps (MSTmap) generation in~\cite{niu2020video}. As shown in Fig.~\ref{fig:framework} (left), We first use an open source face detector OpenFace~\cite{baltrusaitis2018openface} to detect the facial landmarks, based on which we define the most informative region of interest (ROIs) for physiological measurement in face, i.e., the forehead and cheek areas. Specifically, for the $t$-th frame of a video, we first get a set of $n$ informative regions of face $R_{t}=\left \{ R_{1t}, R_{2t},...,R_{nt}\right \}$. Then we calculate the average RGB pixel values of each color channel for all the non-empty subsets of $R_{t}$, which are $2^{n}-1$ combinations of the elements in $R_{t}$. For each video with $T$ frames, the $2^{n}-1$ temporal signals of each channel are placed into rows, and we can get the face MSTmap $MSTmap_{face}$ with the size of $(2^{n}-1) \times T \times 3$ for each video. Besides raw signals from RGB channels, we also consider $MSTmap_{face}$ from YUV color space or other rPPG signal measurement approaches~\cite{wang2014exploiting,wang2016algorithmic}, which will be discussed in Sec.~\ref{sec:ablation}.

Besides the possible skin color changes in facial regions, the environmental patterns from background regions are also helpful for mask PAD. On one hand, the environmental light clues from the background benefit the recovery of the robust facial rPPG signals. On the other hand, the latent noise (e.g., camera noise) from the background can be treated as a contrast with facial signals (large difference for the bonafide while similarity for mask attacks) for liveness discrimination. Therefore, as shown in Fig.~\ref{fig:framework} (left), we also define $m$ background regions on both sides of the face to construct the background MSTmap $MSTmap_{bg}$ with the size $(2^{m}-1) \times T \times 3$ for each video.

\subsection{rPPG Transformer for 3D Mask Face PAD}
\label{Trans}

%As illustrated in Fig.~\ref{fig:framework}, the constructed face and background MSTmaps would be tokenized and then encoded by several shared transformer layers, respectively. Subquently, the token features from face and background are combined and cascaded with an extra transformer layer for context aggregation. Finally, linear classification head is used for mask PAD.  

\noindent\textbf{Image Sequentialization.} \quad  Similar to ViT~\cite{dosovitskiy2020image}, we first preprocess the input face and background MSTmaps into a sequence of flattened patches $X_{face}$ and $X_{bg}$, respectively. However, the original split method cuts the images into non-overlapping patches, which discards discriminative information of some local neighboring signal clips or informative interested regions. To alleviate this problem, we propose to generate overlapping patches with sliding window. Specifically, with the targeted patch size $P_{H}\times P_{W}$ and step size $S_{H}\times S_{W}$ of sliding window, the input face MSTmap with size $H \times W$ will be split into $N$ patches where
\begin{equation} 
N=N_{H}\times N_{W}=\left \lfloor \frac{H-P_{H}+S_{H}}{S_{H}} \right \rfloor \times \left \lfloor \frac{W-P_{W}+S_{W}}{S_{W}} \right \rfloor.
\label{eq:token}
%\vspace{-2.1em}
\end{equation}
 Similarly, the background MSTmap is partitioned into $M$ patches. This simple yet fine-grained patch partition strategy improves the performance remarkably.

\noindent\textbf{Patch Embedding.} \quad We map the vectorized patches $X_{face}$
into a latent $D$-dimensional embedding space using a trainable linear projection. Then, the patch embeddings $Z^{0}_{face}$ can be formulated with an additional learnable position embedding to retain positional information:
\begin{equation} 
Z^{0}_{face}=[X_{faceCls};X_{face}^{1}\mathbf{E},X_{face}^{2}\mathbf{E},...,X_{face}^{N}\mathbf{E}]+\mathbf{E}_{pos},
\label{eq:embedding}
%\vspace{-2.1em}
\end{equation}
where $\mathbf{E}\in \mathbb{R}^{(P^{2}\cdot C)*D}$ is the patch embedding projection from the original $C$-channel space, and $\mathbf{E}_{pos}\in \mathbb{R}^{(N+1)\times D}$ denotes the position embedding. The class token $X_{faceCls}$, i.e., a learnable embedding, is concatenated to the patch
embeddings. Similarly, the background embedding $Z^{0}_{bg}$ can be extracted using the shared linear projection $\mathbf{E}$ but independent class token $X_{bgCls}$ and position embedding $\mathbf{E}^{'}_{pos}$.

\noindent\textbf{Transformer Encoder.} \quad  The Transformer encoder contains $L$ layers of
multihead self-attention (MSA) and multi-layer perceptron (MLP) blocks. Thus the output of the $l$-th layer can be written as follows:
\begin{equation} 
\begin{split}
&Z^{l'}=MSA(LN(Z^{l-1}))+Z^{l-1},  \quad  l\in 1,2,...,L \\
&Z^{l}=MLP(LN(Z^{l'}))+Z^{l'},  \quad  l\in 1,2,...,L
\end{split}
\label{eq:trans}
%\vspace{-2.1em}
\end{equation}
where $Z^{l}\in \mathbb{R}^{(N+1)\times D}$ is the token features in the $l$-th layer. $LN(.)$ denotes the layer normalization operation, and MSA is composed of $h$ parallel self-attention (SA),
\begin{equation} 
\begin{split}
&[Q,K,V]=Z^{l}U_{QKV},\\
&SA(Z^{l})=Softmax(QK^{T}/\sqrt{D^{'}})V,
\end{split}
\label{eq:MSA}
%\vspace{-2.1em}
\end{equation}
where $U_{QKV}\in \mathbb{R}^{D\times3D^{'}}$ is the weight matrix for linear transformation, and $A=Softmax(QK^{T}/\sqrt{D^{'}})$ is the attention map. The output of MSA is the concatenation of $h$ attention head outputs
\begin{equation} 
MSA(Z^{l}) = [SA_{1}(Z^{l}); SA_{2}(Z^{l});...; SA_{h}(Z^{l})]U_{MSA},
\end{equation}
where $U_{MSA}\in \mathbb{R}^{hD^{'}\times D}$. As illustrated in Fig.~\ref{fig:framework}, the two transformer branches (i.e., face and background) first project the patch embeddings $Z^{0}_{face}$ and $Z^{0}_{bg}$, and then $L$-layer shared transformer encoder is utilized for global attentional features ($Z^{L}_{face}$ and $Z^{L}_{bg}$) representation. Subsequently, all token features from $Z^{L}_{face}$ and $Z^{L}_{bg}$ (except the respective class token features) as well as an joint class token $X_{comCls}$ are concatenated, and an extra transformer layer is utilized for face and background context aggregation. 

%\noindent\textbf{Loss Functions.} \quad
\noindent\textbf{Hierarchical Supervision.} \quad  In order to provide explicit supervision signals for TransRPPG, we design three Binary Cross Entropy (BCE) losses (bonafide vs. mask attack) for face, background, and combined branches, respectively. To be specific, three class tokens $X_{faceCls}$, $X_{bgCls}$, and $X_{comCls}$ with independent linear classification heads are supervised with BCE losses $\mathcal{L}_{face}$, $\mathcal{L}_{bg}$, and $\mathcal{L}_{combined}$, respectively. The overall loss can be formulated as $\mathcal{L}_{overall}=\mathcal{L}_{face}+\mathcal{L}_{bg}+\mathcal{L}_{combined}$. As there are usually no liveness clues in the background regions, the groundtruth for $\mathcal{L}_{bg}$ are all simply regarded as `mask attack'.

\begin{figure}
\centering
\includegraphics[scale=0.38]{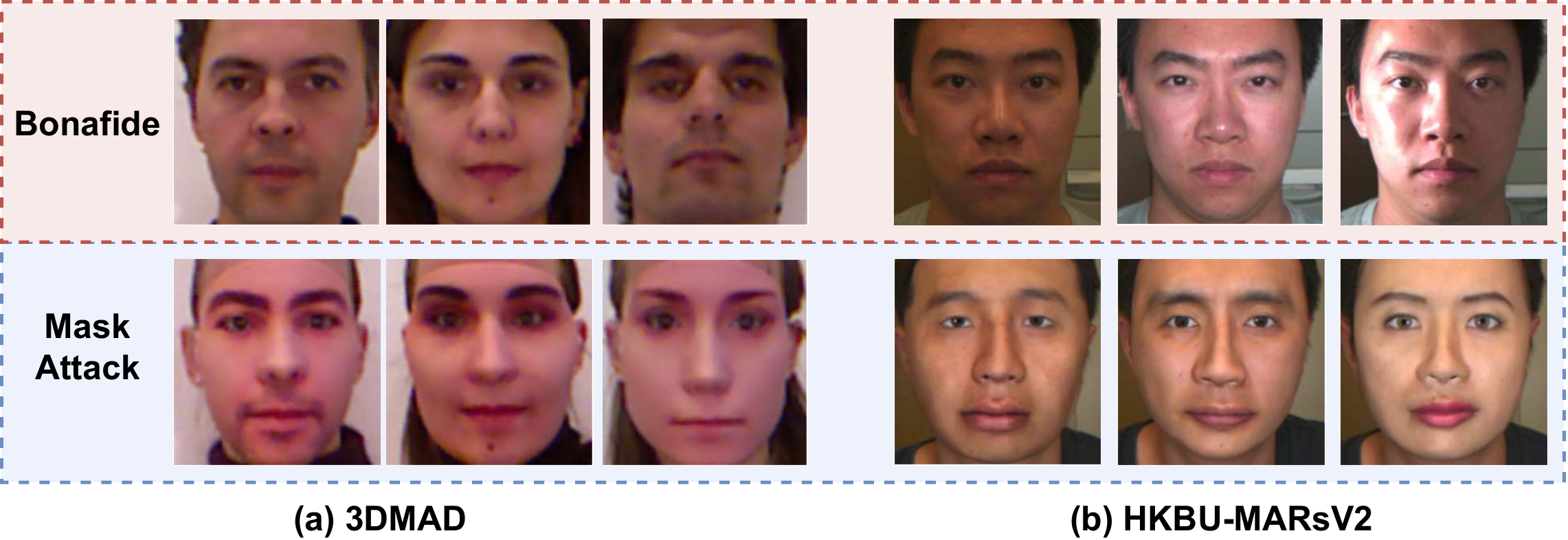}
\vspace{-1.8em}
\caption{\small{Face samples from two 3D mask face PAD datasets: (a) 3DMAD~\cite{erdogmus2014spoofing}, and (b) HKBU-MARsV2~\cite{liu20163d}.}}
\label{fig:dataset}
\vspace{-1.5em}
\end{figure}

\section{Experiments}
\label{sec:guidelines}

\subsection{Datasets and Metrics}
Two public datasets (see Fig.~\ref{fig:dataset}) are employed in our experiments. The \textbf{3DMAD}~\cite{erdogmus2014spoofing} dataset contains 255 videos from 17 subjects with 17 low-fidelity hard resin
masks, which are recorded at 640 × 480, 30fps under controlled lighting condition. The \textbf{HKBU-MARsV2}~\cite{liu20163d} dataset consists of 1008 videos from 12 subjects with 6 low- and 6 high-fidelity hard resin
masks, which are recorded at 1280 × 720,
25fps under variant room lights. In terms of the intra-dataset testings on 3DMAD, MARsV2, and combined dataset (3DMAD+MARsV2), leave-one-subject-out cross-validation protocol is used. As for the cross-dataset testings between 3DMAD and MARsV2, all videos in one dataset are used for training while those in the other one for testing. Area Under the Curve (AUC), Equal Error Rate (EER), Half Total Error Rate (HTER)~\cite{erdogmus2014spoofing}, and False Fake Rate (FFR) when False Liveness Rate (FLR) equals to 0.01 are
used as evaluation metrics.

\begin{table*}[t]\small
\vspace{-0.3em}
\centering
\caption{Intra-dataset results on 3DMAD, MARsV2, and combined (3DMAD+MARsV2) datasets. Best results are marked in \textbf{bold} and second best in \underline{underline}. The rPPG-based methods are marked with $\star$ while appearance-based methods with $\blacktriangle$.} 
\vspace{-0.5em}
\label{tab:IntraResults}
\resizebox{0.92\textwidth}{!} {\begin{tabular}{l| c c c| c  c c |c c c } 
 \toprule[1pt]
 \multirow{2}{*}{Method} & \multicolumn{3}{c|}{3DMAD} & \multicolumn{3}{c|}{MARsV2}   & \multicolumn{3}{c}{Combined} \\
 %\hline
 
 & EER(\%)$\downarrow$ & AUC(\%)$\uparrow$ & FFR@FLR=0.01$\downarrow$  & EER(\%)$\downarrow$ & AUC(\%)$\uparrow$  & FFR@FLR=0.01$\downarrow$  & EER(\%)$\downarrow$ & AUC(\%)$\uparrow$ &  FFR@FLR=0.01$\downarrow$ \\

 \midrule
 MS-LBP~\cite{erdogmus2014spoofing}$\blacktriangle$ & \textbf{2.71} & \textbf{99.7} & \textbf{3.62} &  22.5 & 85.8 & 95.1 & 16.6 & 91.0 & 64.2\\
CTA~\cite{boulkenafet2015face}$\blacktriangle$ & 4.24 & \underline{99.3} & \underline{11.8} &  23.0 & 82.3 & 89.2 & 18.9 & 87.7 & 95.7\\
VGG16~\cite{simonyan2014very}$\blacktriangle$ & 6.63 &  98.9 & 18.5 &  15.2 & 91.4 & 93.5& 14.5 & 93.5 & 71.5\\
 
 \midrule
GrPPG~\cite{li2016generalized}$\star$ & 14.4 &   92.2 &  36.0 &  16.4 & 89.4 & 32.9 & 15.2  & 91.1 & 42.8\\
LrPPG~\cite{liu20163d}$\star$ & 9.64 &  95.5 & 14.8 & 9.07 &  \underline{97.0} & 38.9 & 9.21 & 95.7 & 29.4\\
CFrPPG~\cite{liu2018remote}$\star$ & 7.44 &  96.8 & 13.6 &  \textbf{4.04} &  \textbf{99.3} & \textbf{17.8} & \underline{6.54} & \underline{97.6} & \textbf{15.5}\\
\textbf{TransRPPG (Ours)$\star$} & \underline{2.38} & 98.77 & 13.73 &  \underline{8.47} & 96.82 & \underline{29.79}& \textbf{5.93} & \textbf{97.95} & \underline{22.46}\\
% \textbf{AutoHR (Ours)$\star$} & \underline{8.48
%} & \underline{5.68} & \underline{8.68} & %\underline{0.72}\\

 \bottomrule[1pt]
 \end{tabular}}
\vspace{-1.0em}
\end{table*}

\vspace{-1.0em}
\subsection{Implementation Details}

  All 10-second videos are linearly interpolated into 30 fps to keep the same frame numbers ($T=300$). There are $n=6$ facial regions and $m=4$ background regions for two MSTmaps constructions, which results in $MSTmap_{face}$ and $MSTmap_{bg}$ with size 63$\times$300$\times$3 and 15$\times$300$\times$3, respectively. In terms of transformer settings, patch size $P_{H}=3,P_{W}=30$, step size $S_{H}=1,S_{W}=15$, (hidden) embedding channels $D=D^{'}=96$, heads number = 3, and $L=6$ layers are utilized. All MLP blocks has two fully-connected layers, where feature dimension doubles in the first layer. Our proposed method is implemented with Pytorch. The models are trained with Adam optimizer and the initial learning rate (lr) and weight decay are 1e-4 and 5e-5, respectively. We train models on a single V100 GPU with batchsize 10 for maximum 60 epochs while lr halves in the 45th epoch.

\begin{figure}
\centering
\vspace{-1.0em}
\includegraphics[scale=0.35]{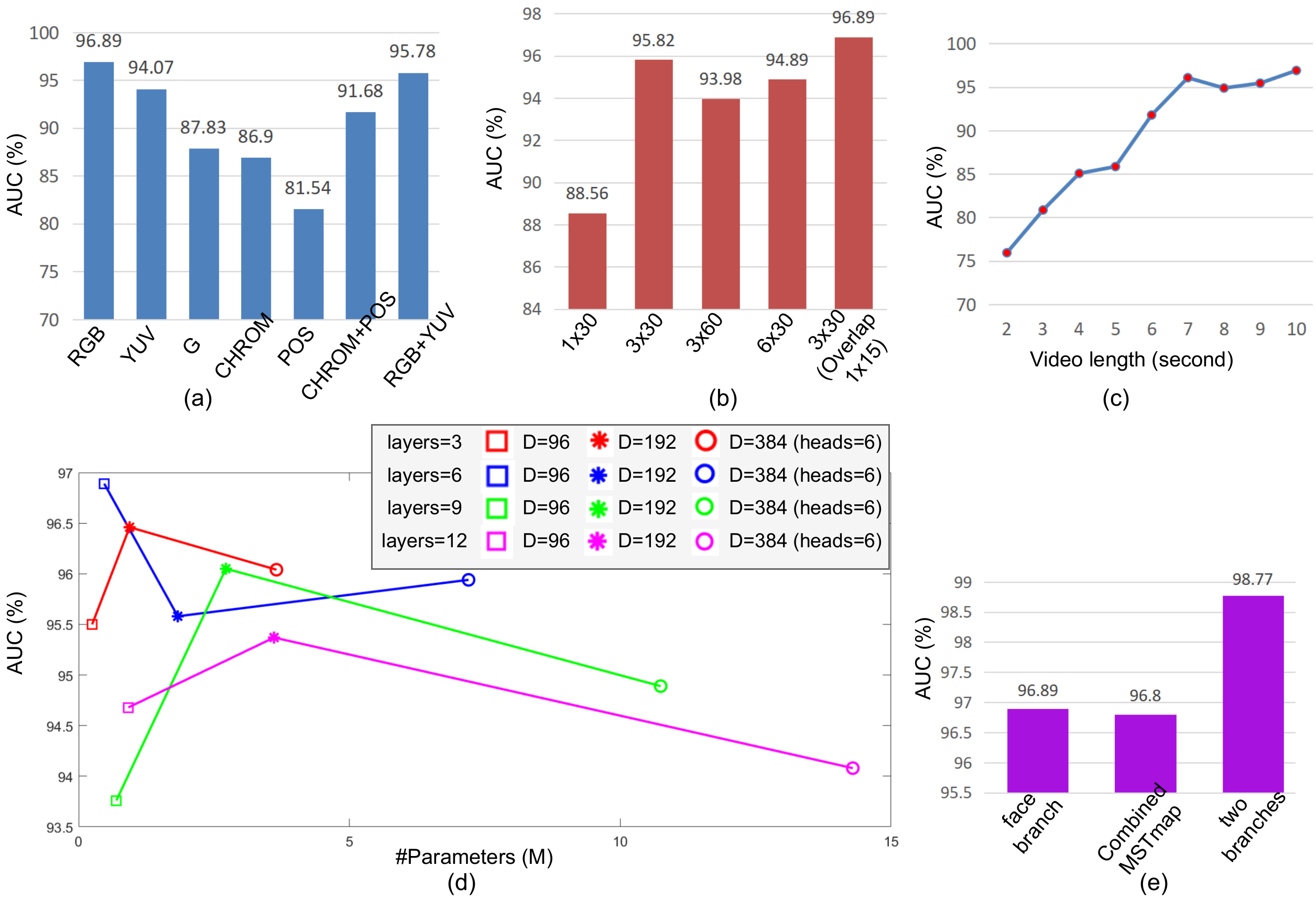}
\vspace{-1.6em}
\caption{\small{Ablation studies on (a) Color space for MSTmap; (b) Patch size for image sequentialization; (c) Video length; (d) Transformers' depth/width; and (e) Background branch.}}
\label{fig:Ablation}
\vspace{-0.7em}
\end{figure}

\vspace{-1.0em}
\subsection{Ablation Study}
\label{sec:ablation}
All ablation studies are intra-dataset tested on the 3DMAD dataset. For convenience, single face branch of TransRPPG is first adopted as the backbone in experiments, and we will study the effects of two-branch framework at last. 

\textbf{Impact of Color Space for MSTmap.}\quad   Fig.~\ref{fig:Ablation}(a) shows that MSTmaps construction from raw RGB channels are best-suited for TransRPPG. Despite great rPPG measurement performance in previous works (e.g., G channel~\cite{li2014remote}, CHROM~\cite{de2013robust}, POS~\cite{wang2016algorithmic}, and RGB+YUV~\cite{niu2020video}), these color spaces cannot provide extra improvement for TransRPPG.

\textbf{Impact of Patch Size.}\quad 
As shown in Fig.~\ref{fig:Ablation}(b), patch size of STMmap tokenization influences liveness representation a lot. When partition with size  1x30 ($P_{H}\times P_{W}$), i.e., only single spatial region clues within each patch, the AUC drops sharply. In contrast, large spatial (e.g., $P_{H}$= 6) or temporal (e.g., $P_{W}$= 60) patch sizes could not bring extra improvements. Patch size with 3x30 performs the best while introducing  patch overlapping 1x15 ($S_{H}\times S_{W}$) sampling boosts 1\% AUC.               

\textbf{Impact of Video Length.}\quad 
Fig.~\ref{fig:Ablation}(c) illustrates that AUC ascends reasonably when video length increases because of richer temporal context for global modeling. It is interesting to find that using 7s video clips could still achieve comparable performance (96.06\% vs 96.89\% AUC) as 10s ones. We also find the limitations that TransRPPG under time-constrained scenarios ($\textless$5s) have poor performance ($\textless$90\% AUC).

\textbf{Impact of Transformers' Depth/Width.}\quad It is necessary to investigate the impact of transformers' depth/width for mask PAD. As illustrated in Fig.~\ref{fig:Ablation}(d), layer number (depth) plays more important role than embedding dimension (width). TransRPPG with shallow layers (e.g., layer = 3 and 6) performs better than those with deeper layers (e.g., layer = 9 and 12). The highest AUC is achieved when $L$ = 6 with $D$ = 92.

\textbf{Effects of the Background Branch.}\quad 
Fig.~\ref{fig:Ablation}(e) illustrates the evaluation results w/ and w/o $MSTmap_{bg}$. There are two configurations with background map: 1) using combined MSTmap via concatenated $MSTmap_{face}$ and $MSTmap_{bg}$ for single face branch input; and 2) two-branch framework in Fig.~\ref{fig:framework}. It is surprised that combined MSTmap performs even worse than $MSTmap_{face}$, which might be caused by the inharmonious attention learning within compound maps. In contrast, disentangled learning from two branches improves the AUC by nearly 2\%. We also find that the constraints $\mathcal{L}_{face}$ and $\mathcal{L}_{bg}$ are helpful for liveness feature learning, improving the AUC from 97.51\% to 98.77\%.

\begin{table}[t]\small
\vspace{-0.5em}
\centering
\caption{Cross-dataset results on 3DMAD and MARsV2. } 
\vspace{-0.3em}
\label{tab:CrossResults}
\resizebox{0.47\textwidth}{!} {\begin{tabular}{l| c c | c c   } 
 \toprule[1pt]
 \multirow{2}{*}{Method} & \multicolumn{2}{c|}{3DMAD$\rightarrow$MARsV2} & \multicolumn{2}{c}{MARsV2$\rightarrow$3DMAD}  \\
 %\hline
 
 &  AUC(\%)$\uparrow$ & FFR@FLR=0.01$\downarrow$  &  AUC(\%)$\uparrow$ & FFR@FLR=0.01$\downarrow$ \\

 \midrule
 MS-LBP~\cite{erdogmus2014spoofing}$\blacktriangle$ & 60.4 & 100.0 & 75.3 &  87.8  \\
 
CTA~\cite{boulkenafet2015face}$\blacktriangle$ & 62.1 & 98.3 & 60.5 &  96.5 \\

VGG16~\cite{simonyan2014very}$\blacktriangle$ & 54.6 &  97.9 & 58.6 &  99.3 \\
 
 \midrule
GrPPG~\cite{li2016generalized}$\star$ &  86.7 &  78.5&  87.2 &  94.5\\

LrPPG~\cite{liu20163d22}$\star$ & \underline{95.6} &  61.7 & 92.3 & 48.7  \\

CFrPPG~\cite{liu2018remote}$\star$ & \textbf{99.0} &  \textbf{19.6}  & \underline{98.0} &  \textbf{12.4}  \\

\textbf{TransRPPG (Ours)$\star$} & 91.3 & \underline{47.6} & \textbf{98.3} &  \underline{18.5} \\

% \textbf{AutoHR (Ours)$\star$} & \underline{8.48
%} & \underline{5.68} & \underline{8.68} & %\underline{0.72}\\

 \bottomrule[1pt]
 \end{tabular}}
\vspace{-1.0em}
\end{table}

\vspace{-0.6em}
\subsection{Intra-dataset Testing}
In this subsection, we compare our TransRPPG with three previous rPPG-based mask PAD methods (GrPPG~\cite{li2016generalized}, LrPPG~\cite{liu20163d22}, and CFrPPG~\cite{liu2018remote}) as well as appearance-based baselines on 3DMAD, MARsV2, and combined datasets. As shown in the upper block of Tab.~\ref{tab:IntraResults}, appearance-based methods perform well on 3DMAD but poorly on MARsV2 because the 3D mask attacks on the latter one are with higher fidelity and harder to distinguish via texture clues. In contrast, four rPPG-based approaches shown in the lower block have consistent good performance on both 3DMAD and MARsV2 datasets, indicating that the learned rPPG-based liveness features are independent from mask fidelity. Instead of LrPPG and CFrPPG using complex spectrum features, our TransRPPG uses simple time-domain signals, and achieves the best EER and AUC on the combined dataset. Introducing frequency-domain representation to TransRPPG might be a possible future direction.

\vspace{-0.6em}
\subsection{Cross-dataset Testing}
 Here we alternatively train and test between 3DMAD and MARsV2 to validate the models' generalization capacity. As shown in Tab.~\ref{tab:CrossResults}, it is clear that the proposed TransRPPG generalizes well from MARv2 to 3DMAD because the training samples in MARv2 are more sufficient and diverse to alleviate the overfitting problem. On the contrary, TransRPPG suffers from obvious performance drops when trained on 3DMAD and tested on MARsV2 due to the small amount of training data under single scenario. LrPPG and CFrPPG generalize well in cross testings, implying the importance of spectrum representation lacked in TransRPPG.

\vspace{-0.4em}
\section{Conclusion}
\vspace{-0.1em}

In this letter, we propose a lightweight remote photoplethysmography transformer (TransRPPG) for 3D mask face presentation attack (PA) detection based on the facial and background multiscale spatio-temporal maps. In the future, we plan to explore TransRPPG on 1) detecting other PA types such as print, replay, and makeup; and 2) various rPPG-based applications like heart rate and stress estimation.

\bibliographystyle{IEEEtran}
% argument is your BibTeX string definitions and bibliography database(s)
\bibliography{IEEEabrv,reference}

\clearpage

\thispagestyle{empty}

\appendix

\section{Appendix A: Visualization}

\begin{figure*}[bp]
\centering
\includegraphics[scale=0.6]{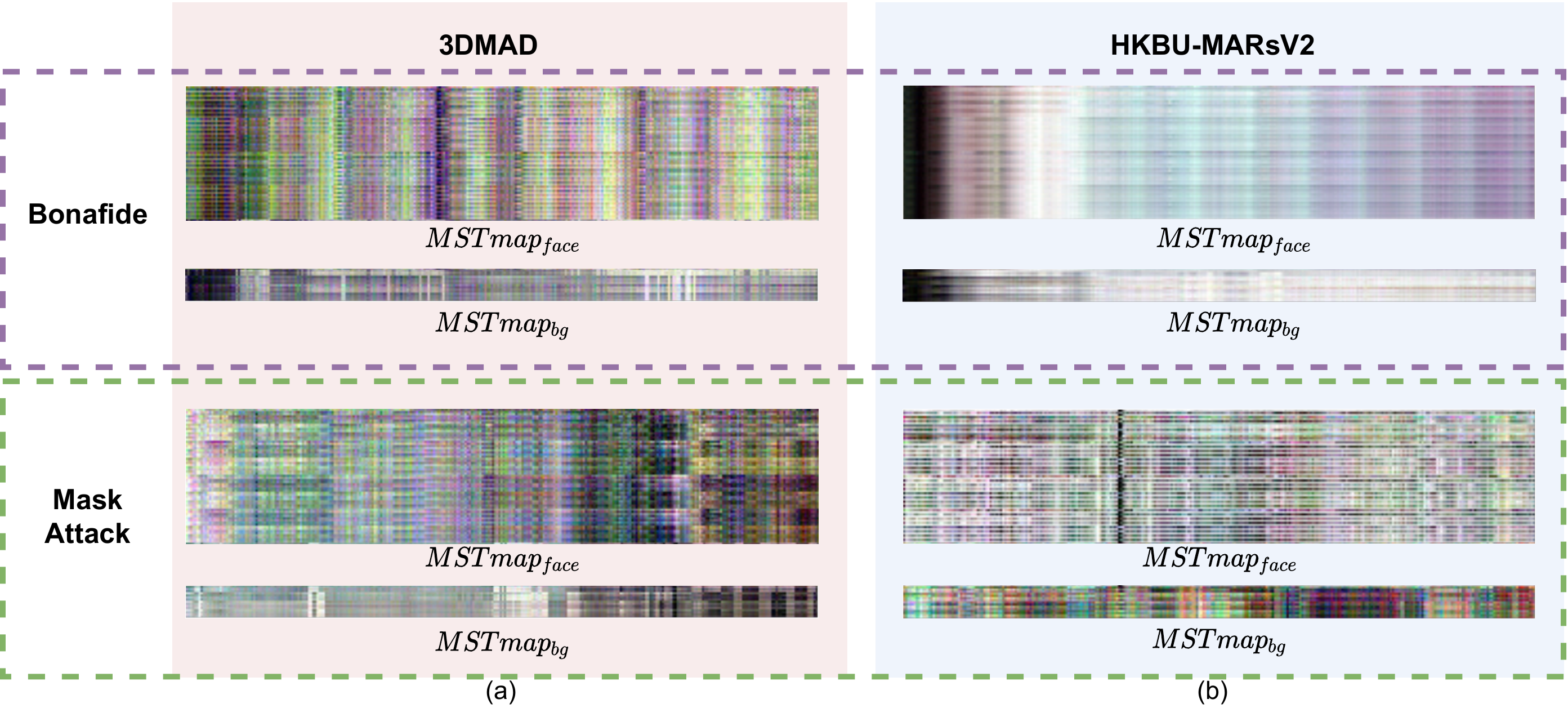}
\vspace{-0.5em}
\caption{\small{Visualization of MSTmaps on (a) 3DMAD, and (b) MARsV2. }}
\label{fig:visualization1}
\end{figure*}

\textbf{Visualization of MSTmap.}\quad 
Here some typical MSTmaps from 3DMAD and MARsV2 datasets are shown in Fig.~\ref{fig:visualization1}. It is obvious that face MSTmaps from the bonafide are more periodic and regular compared with those from the 3D mask attacks. Moreover, the background MSTmaps from both bonafide and 3D masks are relatively noisy and nonperiodic.

\begin{figure*}[bp]
\centering
\includegraphics[scale=0.56]{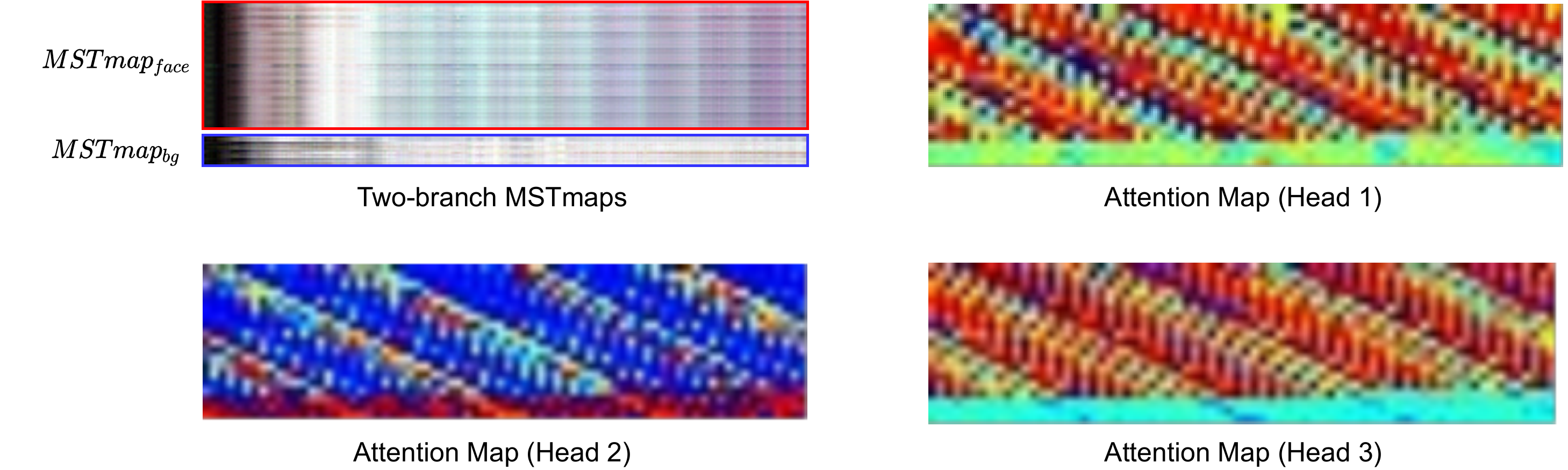}
\vspace{-0.5em}
\caption{\small{Visualization of the attention maps in the last transformer layer on a bonafide sample of MARsV2 dataest.}}
\label{fig:visualization2}
\end{figure*}

\textbf{Attention Maps in TransRPPG.}\quad 
We visualize the attention maps of the last transformer layer from the trained TransRPPG in Fig.~\ref{fig:visualization2}. Given the facial and background MSTmaps from the bonafide input, token features $Z^{L}_{face}$ and $Z^{L}_{bg}$ are first extracted from face branch and background branch, respectively. Then the concatenated facial and background features are further refined via an extra (last) transformer. It can be seen from Fig.~\ref{fig:visualization2} that attention maps from the bonafide sample share the similar attentions on all three heads, i.e., periodic cross activations in the facial patches while ignoring the noisy interference from background regions. In other words, TranRPPG is able to find the global periodicity across the rPPG signals from different facial regions, which is promising for various rPPG-based tasks such as average heart rate estimation.

\section{Appendix B: Ablations on Class Token and Position Embedding}
\textbf{Effect of Class Token.}\quad 
We also investigate the effect of class tokens in TransRPPG. An alternative solution for global feature aggregation is to use global average pooling (GAP) on the top of transformer layers before classification heads. However, TransRPPG with class tokens outperforms that with GAP remarkably (98.77\% vs 96.37\% AUC) on 3DMAD dataset, indicating the effectiveness of class tokens. 

\textbf{Effect of Position Embedding.}\quad 
Position embedding is proved to be vital for vision transformer because of its spatial localization knowledge. Here we conduct an ablation to verify the importance of position embedding for TransRPPG. Specially, we find that when removing position embedding, the AUC is decreased by 1.62\% on 3DMAD dataset. In other words, the learnable position embedding provides extra positional inductive bias for better token interaction.

\end{document}